\documentclass[runningheads]{llncs}
\pdfoutput=1

\usepackage{graphicx}
\usepackage{amsfonts}

\usepackage{adjustbox}

\usepackage{hyperref}

\usepackage{amsmath}
\usepackage{breqn}

\newcommand{\etal}{\textit{et al}.}

\begin{document}
\title{TAPE: Temporal Attention-based Probabilistic human pose and shape Estimation}
\titlerunning{Temporal Attention-based Probabilistic human pose and shape Estimation}
%

\author{Nikolaos Vasilikopoulos\inst{1,2}\orcidID{0000-0002-2340-8039}\and
Nikos Kolotouros\inst{}\orcidID{0000-0003-4885-4876} \and
Aggeliki Tsoli\inst{2}\orcidID{0000-0002-7254-3747}  \and \\
Antonis Argyros\inst{1,2}\orcidID{0000-0001-8230-3192}}
\authorrunning{Nikolaos Vasilikopoulos et al.}
%
\institute{Computer Science Department, University of Crete \\
 \and
Foundation for Research and Technology - Hellas (FORTH)\\
\email{\{nvasilik,aggeliki,argyros\}@ics.forth.gr}\\
\email{nikoskolot@gmail.com}}


\maketitle             
\begin{abstract}
Reconstructing 3D human pose and shape from monocular videos is a well-studied but challenging problem. Common challenges include occlusions, the inherent ambiguities in the 2D to 3D mapping and the computational complexity of video processing. Existing methods ignore the ambiguities of the reconstruction and provide a single deterministic estimate for the 3D pose. In order to address these issues, we present a  \textbf{T}emporal \textbf{A}ttention based \textbf{P}robabilistic human pose and shape \textbf{E}stimation method (TAPE) that operates on an RGB video. More specifically, we propose to use a neural network to encode video frames to temporal features using an attention-based neural network. Given these features, we output a per-frame but \textit{temporally-informed} probability distribution for the human pose using Normalizing Flows. We show that TAPE outperforms state-of-the-art methods in standard benchmarks and serves as an effective video-based prior for optimization-based human pose and shape estimation. Code is available at: \url{https://github.com/nikosvasilik/TAPE}

\keywords{3D human pose \and human shape \and normalizing flows \and probabilistic \and  human body reconstruction}

\end{abstract}

\section{Introduction}
\label{sec:intro}
Human pose and shape estimation from RGB video is a key problem in computer vision with a wide variety of applications, such as AR/VR, surveillance, human-robot interaction and more. This problem is particularly challenging due to the large number of Degrees of Freedom of the human body in terms of pose and shape, the self-occlusions among body parts and the inherent ambiguity in 3D skeletal pose estimation given only 2D video observations.
Moreover, processing a video sequence requires increased computational resources compared to single frame processing.  
\begin{figure}[t]
    \centering
    \includegraphics[width=0.7\textwidth]{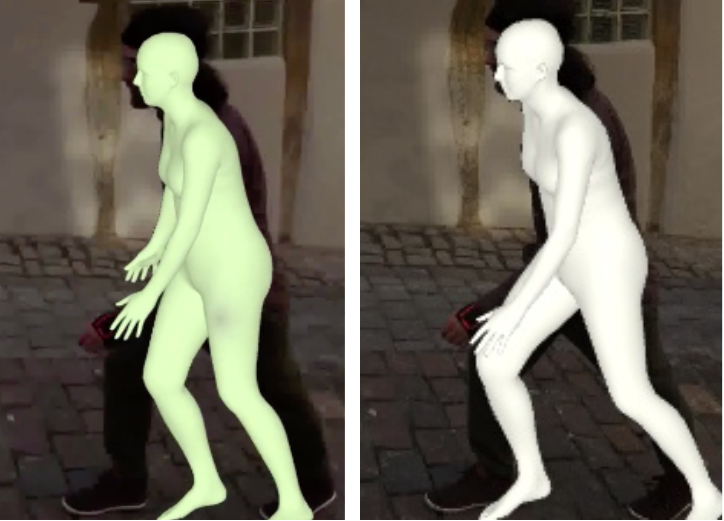}
    \caption{Left (Green): 3D body pose and shape estimation with MPS-Net~\cite{WeiLin2022mpsnet}. Right (White): 3D body pose and shape estimation with the proposed method (TAPE). As it can be verified, the 3D human shape  and pose estimated by TAPE is in better agreement with the shape and pose of the imaged person. }
    \label{mps-TAPE}
\end{figure}

State-of-the-art previous work relies on deep learning and has mostly approached human pose and shape estimation from video as independent estimation of human pose and shape from a single RGB image~\cite{kolotouros2019spin,kolotouros2019cmr,kanazawaHMR18,Bogo:ECCV:2016,omran2018neural}. However, the limited previous efforts that take into account the temporal aspect of a video~\cite{choi2020beyond,kocabas2019vibe,humanMotionKanazawa19} have shown increased accuracy and temporal coherence compared to methods that operate using a single RGB image as input.
Most of the datasets used for training have only 2D annotations for the human body joints~\cite{6751390,humanMotionKZFM19} and datasets with 3D annotations are mostly captured in controlled lab settings~\cite{h36m_pami}.
Contrary to the vast majority of existing methods that provide deterministic outputs of human pose and shape from a single RGB image~\cite{kolotouros2019spin,kolotouros2019cmr,kanazawaHMR18,Bogo:ECCV:2016}, recent efforts for probabilistic human pose estimation~\cite{kolotouros2021prohmr} have shown increased performance due to dealing more effectively with the 2D-3D ambiguity of the observations in existing datasets.      

In this paper, we propose the first temporal and probabilistic method for human pose and shape estimation from video. The human body is represented using the widely used SMPL model~\cite{SMPL:2015}. We propose a deep learning architecture where static ResNet-50~\cite{7780459} features are extracted for each frame in the video and are then converted to temporal features using an attention-based temporal encoder and then integrated to one temporal feature as proposed by Wei~\etal~\cite{WeiLin2022mpsnet}. The output is a probability distribution for the human pose using Normalizing Flows
and point estimates for the human shape and camera parameters inspired by Kolotouros et al~\cite{kolotouros2021prohmr}. Extensive experimental results on well-established datasets show increased accuracy for the task of regressing human pose and shape from visual data using a neural network. In addition, we demonstrate that our work serves as an effective video-based prior for optimization-based human prediction.  
In summary, our contributions are the following:
 \begin{itemize}
     \item We propose a temporal probabilistic model for human body and shape estimation from video input.
     
     \item We extend a state-of-the-art optimization method for human model fitting to 2D observations using our model as a video-based prior. 
     
     \item We show state-of-the-art 3D pose estimation in standard benchmarks.

 \end{itemize}
 \section{Related Work}
\subsection{Human 3D shape and pose from a single RGB image}

\vspace*{0.1cm}\noindent\textbf{Regression}: Most of previous work on human pose and shape estimation from video input has treated the problem as human prediction from a single image handling each frame independently. These methods typically follow the regression paradigm where the parameters of a parametric model~\cite{SMPL:2015,SMPL-X:2019} are regressed from a deep neural network, given a single image as input ~\cite{guler2019holopose,kolotouros2019convolutional,pavlakos2019texturepose,choutas2020monocular,georgakis2020hierarchical,jiang2020coherent}.
A representative and baseline method is HMR~\cite{kanazawaHMR18}, which regresses SMPL~\cite{SMPL:2015} parameters by minimizing the 3D to 2D keypoint reprojection loss while also using a pose discriminator for adversarial training. Similarly to HMR, we use a multi-layer perceptron for estimating human shape and camera parameters.\\

\vspace*{0.1cm}\noindent\textbf{Optimization}: Optimization-based methods estimate iteratively the parameters of a body model, such that it is consistent with a set of 2D cues. Several cues have been employed, e.g., silhouettes~\cite{lassner2017unite}, POFs~\cite{xiang2019monocular}, dense correspondences~\cite{guler2019holopose} or contact~\cite{muller2021self}. The most widely used cues are 2D keypoints with SMPLify~\cite{Bogo:ECCV:2016} being a representative optimization-based approach which predicts SMPL parameters from 2D keypoints. \\

\vspace*{0.1cm}\noindent\textbf{Hybrid Optimization-Regression}: Optimization-based approaches are slower than regression-based ones, but can be more accurate given a good initialization. Thus, it is a common practice that a regression-based method is used to initialize an optimization-based method. SPIN~\cite{kolotouros2019spin} is a hybrid method which uses regression and optimization in the training loop. EFT~\cite{joo2020eft} combines optimization and regression but also updates the network weights during the fitting procedure.
In this work, we demonstrate how our probabilistic model can leverage video-based information to effectively guide keypoint-based optimization. \\

\vspace*{0.1cm}\noindent\textbf{Non Parametric}:
METRO~\cite{lin2021end-to-end} and Mesh Graphormer~\cite{lin2021-mesh-graphormer} regress the 3D mesh vertices of the human body and do not predict the model parameters, directly. Also, they use the HRNet~\cite{Zhou_2019_CVPR} backbone instead of ResNet~\cite{7780459}. Therefore, despite their SOTA performance, these methods cannot be directly compared with our method and the methods we compare with.

\subsection{Human 3D shape and pose from RGB video}
Most related to our work are methods for human pose and shape prediction leveraging the temporal aspect of a video. Kanazawa \etal~\cite{humanMotionKanazawa19} proposed a regression-based method to learn human motion kinematics by predicting past and future frames with a temporal encoder.
Kocabas \etal~\cite{kocabas2019vibe} proposed VIBE, a temporal encoder with Gated Recurrent Units (GRUs) to capture the motion between the static features and a motion discriminator trained with the AMASS~\cite{AMASS:ICCV:2019} dataset for adversarial training. TCMR~\cite{choi2020beyond} reduced drastically the acceleration error of VIBE by using a temporal encoder which consists of 3 GRUs,  one for the past, one for the future and one for the current frame and then integrating the features to produce a smooth video result with better pose and acceleration estimation than previous works.
Recently, MPS-Net~\cite{WeiLin2022mpsnet} proposed a motion continuity attention (MoCA) module which captures the continuity between frames and a hierarchical attentive feature integration (HAFI) module to effectively combine adjacent past and future feature representations to strengthen temporal correlation and refine the feature representation of the current frame. MPS-Net achieves temporally coherent pose estimates without penalizing the accuracy on pose prediction. Our method adopts the MoCA and the HAFI modules from MPS-Net as well as the motion discriminator in VIBE.

\subsection{Multiple hypotheses}
Biggs \etal~\cite{biggs2020multibodies} extend HMR~\cite{kanazawaHMR18} with $N$ prediction heads. This leads to a discrete set of hypotheses, instead of a full probability of poses as we do. In a concurrent work, Sengupta \etal~\cite{sengupta2021hierprobhuman} use a Gaussian posterior to model the uncertainty in
the parameter prediction. 
Recently, ProHMR~\cite{kolotouros2021prohmr} used Normalizing Flows
to predict a  distribution of 3D poses conditioned on the provided 2D input. The probabilistic modeling of ProHMR is efficient at computing the most likely pose comparable to the SOTA methods and also outperforms previous work on optimization tasks such as 2D keypoint fitting and multi-view refinement. Our method uses the probabilistic model of ProHMR for video input.
\section{Method}
\begin{figure*}[t]
    \centering
    \includegraphics[width=1.0\linewidth]{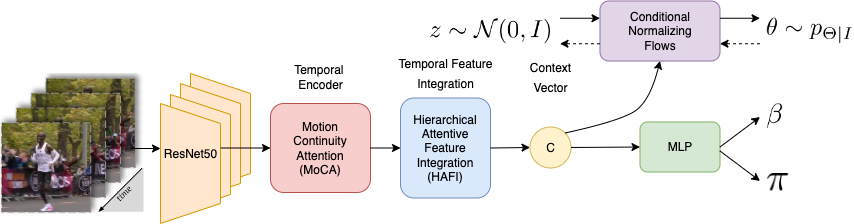}
    \caption{TAPE architecture: We extract static ResNet-50 features from video which are used as input for the Temporal Encoder (MoCA). MoCA outputs temporal features, and the HAFI module integrates them to a single feature, the hidden vector $c$, which is used as the conditioning input to the flow model. In parallel, it is also decoded to shape parameters $\beta$ and camera parameters $\pi$. Our flow model learns an invertible mapping which allows for two processing directions; depending on the desired function, we can perform both sampling and fast likelihood computation.}
    \label{architecture}
\end{figure*}

Given an input video $V=\left\{I_{t}\right\}_{t=1}^{T}$ of length $T$, we resize each frame at a resolution of $224\times224$. The output of TAPE when applied to this input is a per-frame probability distribution for the human pose and point estimates for the human shape as well as camera parameters. We represent the human body using the SMPL~\cite{SMPL:2015} model. 
SMPL provides a function
$ M( \theta , \beta )$ that takes as input the pose parameters $ \theta \in \mathbb{R}^{72}$ and the
shape parameters $ \beta \in \mathbb{R}^{10} $, and returns the body mesh $M \in \mathbb{R}^{N \times 3}$,
with $N = 6890$ vertices.
The proposed deep learning architecture is illustrated in Figure~\ref{architecture} and detailed in the following sections. 

\subsection{Temporal encoding}
\label{sec:temporalencoder}
We use a pretrained ResNet-50~\cite{7780459} backbone from ProHMR that was trained on standard human pose and shape estimation datasets. The backbone extracts static features for each frame of the video. We consider $T=16$ frames at a time and, following Wei~\etal~\cite{WeiLin2022mpsnet}, 
the features are sent to the Motion Continuity Attention module that outputs $T$ temporal features. The $T$ temporal features are integrated to a single feature context vector $c$ by the Hierarchical Feature Integration module proposed by Wei~\etal~\cite{WeiLin2022mpsnet}.

\subsection{Normalizing flows}
Following Kolotouros \etal~\cite{kolotouros2021prohmr},
we model $p (\theta|I)$
using Conditional Normalizing Flows. We learn a mapping
$f : \mathbb{R}^{d} \times \mathbb{R}^{c} \to \mathbb{R}^{d}$
that is bijective in the latent variable $z$ and the pose parameters $\theta$, and is parametrized by the image features $g = c(I)$. More specifically, 
$\theta=f(z; c)$ and $z = f^{-1}(\theta; c)$.
The choice of this particular model class enables to perform both fast
likelihood computation and sampling from the distribution. Another useful property is that, as shown in Kolotouros \etal~\cite{kolotouros2021prohmr}, the mode of the output distribution is the transformation on $z = 0$, i.e.,
\begin{equation}
\theta^{\overset{*}{}}= argmax_{\theta }  (\theta|c) = f(0; c),
\end{equation}
which means that in the absence of additional evidence, the probabilistic model can be used to make predictions by choosing the sample with the maximum probability.

\subsection{MLP}

Camera parameters $\pi$ and SMPL shape parameters $\beta$ are regressed by a small pretrained MLP as in ProHMR. The MLP takes as input the context vector $c$ and outputs matrices of shape and camera parameters per frame.

\subsection{Motion discriminator}
\label{sec:md}
We train a motion discriminator as in Kocabas \etal~\cite{kocabas2019vibe} using the AMASS dataset~\cite{AMASS:ICCV:2019}. The motion discriminator enforces the generator, i.e. our network, to produce plausible human motions and shapes. The discriminator takes as input the poses $\Theta$ that the generator predicted and outputs a value $\in [0,1]$ representing the probability that $\hat{\Theta}$ belongs to the manifold of plausible human motions.
The motion discriminator consists of a GRU with 2 layers and hidden size 1024. The aggregation of hidden states is done by a self-attention mechanism.
The objective function for training the Motion Discriminator is:
\begin{equation}
L_{DM} = E_{\Theta~p_R} [(D_{M}( \Theta ) - 1)^{2}] + E_{\Theta~p_{G}} [D_{M}( \hat{\Theta} )^{2}],
\end{equation}
where $p_R$ is a real motion sequence from the AMASS
dataset, while $p_{G}$ is a generated motion sequence. Since
$D_M$ is trained on ground-truth poses, it also learns plausible body pose configurations. Therefore, the pose discriminator that was used in ProHMR is not needed in our method.

\subsection{Training objective}

Due to lack of ground truth SMPL annotations in some of the datasets, we use mixed training following ProHMR. When SMPL data are available, we minimize the negative log-likelihood of the ground truth examples $ \theta_{gt}$
\begin{equation}
L_{nll} = -\ln{p_{\Theta|V}}(\theta_{gt}|c).
\end{equation}
In every dataset, we minimize the reprojection loss jointly with an adversarial motion prior based on the motion discriminator described in Section~\ref{sec:md}. In notation, 
\begin{equation}
    L_{exp} = E_{z~p_z}[L_{2D}(f(z;c),\beta,\pi)+L_{adv}(f(z;c),]
\end{equation}
where
\begin{equation}
    L_{adv} = E_{ \Theta~p_{G}} [( D_{M}( \hat{ \Theta } ) - 1)^{2}].
\end{equation}

\noindent For each image $I$, the mode $\theta^{*}_{I}$
of the output distribution corresponds to the transformation of $z = 0$. We do this by explicitly supervising $\theta^{*}_I$
with all the available annotations as in a standard regression framework and minimize
\begin{equation}
 L_{mode}=L_{3D}(\theta^{*}_{I},\beta_+L_{2D})+L_{adv}(\theta^{*}_{I}).
\end{equation}
 
 Following ProHMR, we use 6D representation~\cite{Zhou_2019_CVPR} to model rotations and the $L_{orth}$ loss to force the 6D representations of the samples drawn from the distribution to be close
to the orthonormal 6D representation.

 The final training objective becomes:
\begin{equation}
\begin{split}
    L & =\lambda_{nll}L_{nll}\\
      & +\lambda_{exp,2D}L_{exp,2D}+\lambda_{exp,adv}L_{exp,adv}\\
      & +\lambda_{mode,2D}L_{mode,2D}+\lambda_{mode,adv}L_{mode,adv}\\
      & +\lambda_{mode,\theta}L_{mode,\theta}+\lambda_{mode,\beta}L_{mode,\beta}\\
      & +\lambda_{mode,3D}L_{mode,3D}+\lambda_{orth}L_{orth}. \\ 
\end{split}
\end{equation}
The loss weights are defined as: 
$\lambda_{nll} = 0.001$, $ \lambda_{exp,2D} = 0.001$, $ \lambda_{exp,adv} =
\lambda_{mode,adv} = 0.01$, $ \lambda_{mode,2D} = 0.01$, $ \lambda_{mode,3D} = 0.05$,
$ \lambda_{mode,\theta} = 0.001$, $ \lambda_{mode,\beta} = 0.0005$ and $ \lambda_{orth} = 0.1$.
\subsection{Optimization-based model fitting}

Extending the model fitting procedure in ProHMR to the temporal domain, we use a video-based pose prior that models the likelihood of the pose at a specific frame conditioned on the video evidence:
\begin{equation}
E_{\theta|V} = - \ln{ p_{\Theta|V} (\theta|c)}.
\end{equation} 
As initialization for the fitting, we use the mode $\theta^{*}_{I}$ of the conditional distribution calculated from the regression step.

Following SMPLify~\cite{Bogo:ECCV:2016}, $E_J$ penalizes the weighted 2D distance between the projected model joints and the detected joints and $E_\beta$ is a quadratic penalty on the shape coefficients. \\

\noindent The final objective function for the optimization is:
\begin{equation}
E = \lambda_JE_J - \lambda_{V}E_{\theta|V} + \lambda_{\beta}E_\beta.
\end{equation}

\section{Experiments}
\subsection{Training procedure}
The sequence length during training is $T=16$ frames with a mini batch size of 32. Each frame gets augmented as described in Section~\ref{sec:temporalencoder} and then static features are computed using ResNet-50. We initialize our network with pretrained versions of the ResNet-50 network as well as the Normalizing Flows based on the ProHMR checkpoint that is publicly available. We keep the ResNet-50 features fixed, but continue training the Normalizing Flows. During training, we draw 2 samples from the distribution and not only the mode. The motion discriminator takes as input the predicted parameters $\Theta$  with the ground-truth data from AMASS and is trained to predict a single fake/real probability for each sample. We train our network as well as the motion discriminator using Adam optimizers with learning rate equal to $5e-5$.
Training requires at least $7$ epochs and takes about $1$ hour on a single NVidia GTX1080Ti GPU.

\subsection {Datasets}

We train and evaluate our network for human prediction using the following datasets:
\begin{enumerate}
    \item MPI-INF-3DHP~\cite{mono-3dhp2017}: This dataset contains videos of human motion, mostly indoors, at 30fps. MPI-INF-3DHP is augmented with SMPL parameters every 10 frames, derived from SPIN. 
    \item Human3.6M~\cite{h36m_pami,IonescuSminchisescu11}: Human3.6M is a large-scale dataset captured indoors with an optical motion capture system and the training set consists of 8 different subjects performing various actions. Human3.6M was captured at 50fps and we subsample it to 25fps to match MPI-INF-3DHP. We optionally use SMPL annotations acquired with Mosh~\cite{Loper:SIGASIA:2014}.
    \item 3DPW~\cite{vonMarcard2018}: 3DPW was generated using IMU sensors combined with a 2D pose detector to compute ground truth SMPL parameters. We keep the initial resolution of 3DPW at 30fps. 3DPW is the only dataset we use that consists only of outdoor videos.
\end{enumerate}

\subsection{Metrics}
We report Procrustes-Aligned Mean Per Joint
Position Error (PA-MPJPE) to evaluate the accuracy of the obtained 3D poses. Mean Per Joint Position Error (MPJPE) takes also into consideration the global translation/orientation of the human and the predicted camera parameters. Mean Per
Vertex Error (MPVE) is only available in 3DPW and it helps to measure the accuracy in shape prediction. 
We compare TAPE with state-of-the-art single-image and temporal methods. Acceleration error (ACCEL-ERR measured in $mm/s^2
$), calculated as the difference in acceleration between the ground-truth and predicted 3D joints can be important in video methods since it evaluates how temporal coherent the prediction between consecutive frames is. Given though that existing datasets contain limited variation in acceleration, we rely more heavily in our evaluation on the rest of the error metrics.
\begin{table*}[t]
\begin{adjustbox}{width={\textwidth},totalheight={\textheight},keepaspectratio}%

\begin{tabular}{|c|cccc|ccc|ccc|}
\hline
{Models} & \multicolumn{4}{c|}{3DPW}                                                                                                    & \multicolumn{3}{c|}{MPI-INF-3DHP}                                                   & \multicolumn{3}{c|}{Human3.6M}                                                   \\ \cline{2-11} 
                        & \multicolumn{1}{c|}{PA-}      & \multicolumn{1}{c|}{MPJPE}         & \multicolumn{1}{c|}{MPVE}           & ACCEL    & \multicolumn{1}{c|}{PA-}      & \multicolumn{1}{c|}{MPJPE}         & ACCEL & \multicolumn{1}{c|}{PA-}      & \multicolumn{1}{c|}{MPJPE}         & ACCEL \\ 
                        & \multicolumn{1}{c|}{MPJPE}      & \multicolumn{1}{c|}{ }         & \multicolumn{1}{c|}{ }           & -ERR    & \multicolumn{1}{c|}{MPJPE}      & \multicolumn{1}{c|}{ }         & -ERR & \multicolumn{1}{c|}{MPJPE}      & \multicolumn{1}{c|}{}         & -ERR \\ \hline
SPIN                    & \multicolumn{1}{c|}{59.2}          & \multicolumn{1}{c|}{96.9}          & \multicolumn{1}{c|}{116.4}          & 29.8         & \multicolumn{1}{c|}{67.5}          & \multicolumn{1}{c|}{105.0}         & -         & \multicolumn{1}{c|}{41.1}          & \multicolumn{1}{c|}{-}             & 18.3
\\ \hline
ProHMR                    & \multicolumn{1}{c|}{59.8}          & \multicolumn{1}{c|}{-}          & \multicolumn{1}{c|}{-}          & -         & \multicolumn{1}{c|}{65.0}          & \multicolumn{1}{c|}{-}         & -         & \multicolumn{1}{c|}{41.2}          & \multicolumn{1}{c|}{-}             & -      \\ \hline
Biggs                    & \multicolumn{1}{c|}{59.9}          & \multicolumn{1}{c|}{-}          & \multicolumn{1}{c|}{-}          & -         & \multicolumn{1}{c|}{-}          & \multicolumn{1}{c|}{-}         & -         & \multicolumn{1}{c|}{41.6}          & \multicolumn{1}{c|}{-}             & -      \\ \hline
VIBE                    & \multicolumn{1}{c|}{56.5}          & \multicolumn{1}{c|}{93.5}          & \multicolumn{1}{c|}{113.4}          & 27.1         & \multicolumn{1}{c|}{63.4}          & \multicolumn{1}{c|}{97.7}          & 29.0      & \multicolumn{1}{c|}{41.5}          & \multicolumn{1}{c|}{65.9}          & 18.3      \\ \hline
TCMR                    & \multicolumn{1}{c|}{55.8}          & \multicolumn{1}{c|}{95.0}          & \multicolumn{1}{c|}{111.5}          & \textbf{6.7} & \multicolumn{1}{c|}{62.8}          & \multicolumn{1}{c|}{96.5}          & \textbf{9.5}       & \multicolumn{1}{c|}{41.1}          & \multicolumn{1}{c|}{62.3}          & \textbf{5.3}       \\ \hline
MPS-NET                 & \multicolumn{1}{c|}{\textbf{54.0}} & \multicolumn{1}{c|}{91.6}          & \multicolumn{1}{c|}{\textbf{109.6}}          & 7.5          & \multicolumn{1}{c|}{-}             & \multicolumn{1}{c|}{-}             & -         & \multicolumn{1}{c|}{-}             & \multicolumn{1}{c|}{-}             & -         \\ \hline
\textbf{TAPE (Ours)}                 & \multicolumn{1}{c|}{56.6}          & \multicolumn{1}{c|}{\textbf{89.3}} & \multicolumn{1}{c|}{112.5} & 10.7         & \multicolumn{1}{c|}{\textbf{56.7}} & \multicolumn{1}{c|}{\textbf{94.0}} & 12.4      & \multicolumn{1}{c|}{\textbf{39.5}} & \multicolumn{1}{c|}{\textbf{60.0}} & 6.5      \\ \hline
\end{tabular}
\end{adjustbox}
\caption{Evaluation of state-of-the-art single image-based and
video-based methods on the 3DPW, Human3.6M, and MPI-INF-3DHP datasets. Training has been performed on the MPI-INF-3DHP, Human3.6M datasets (not 3DPW).}
\label{wo3dpw}

\end{table*}

\begin{table*}[t]
\begin{adjustbox}{width={\textwidth},,totalheight={\textheight},keepaspectratio}%

\begin{tabular}{|c|cccc|ccc|ccc|}
\hline
{Models} & \multicolumn{4}{c|}{3DPW}                                                                                                   & \multicolumn{3}{c|}{MPI-INF-3DHP}                                                      & \multicolumn{3}{c|}{Human3.6M}                                                 \\ \cline{2-11} 
                        & \multicolumn{1}{c|}{PA-}      & \multicolumn{1}{c|}{MPJPE}         & \multicolumn{1}{c|}{MPVE}           & ACCEL    & \multicolumn{1}{c|}{PA-}      & \multicolumn{1}{c|}{MPJPE}         & ACCEL & \multicolumn{1}{c|}{PA-}      & \multicolumn{1}{c|}{MPJPE}         & ACCEL \\ 
                        & \multicolumn{1}{c|}{MPJPE}      & \multicolumn{1}{c|}{ }         & \multicolumn{1}{c|}{ }           & -ERR    & \multicolumn{1}{c|}{MPJPE}      & \multicolumn{1}{c|}{ }         & -ERR & \multicolumn{1}{c|}{MPJPE}      & \multicolumn{1}{c|}{}         & -ERR \\ \hline
VIBE                    & \multicolumn{1}{c|}{57.7}          & \multicolumn{1}{c|}{91.9}          & \multicolumn{1}{c|}{-}             & 27.1         & \multicolumn{1}{c|}{68.9}          & \multicolumn{1}{c|}{103.9}         & 27.3         & \multicolumn{1}{c|}{53.3}          & \multicolumn{1}{c|}{78.0}          & 27.3         \\ \hline
TCMR                    & \multicolumn{1}{c|}{52.4}          & \multicolumn{1}{c|}{86.5}          & \multicolumn{1}{c|}{103.2}         & \textbf{6.8} & \multicolumn{1}{c|}{63.5}          & \multicolumn{1}{c|}{97.6}          & \textbf{8.5} & \multicolumn{1}{c|}{52.0}          & \multicolumn{1}{c|}{76.0}          & 15.3         \\ \hline
MPS-NET                 & \multicolumn{1}{c|}{52.1}          & \multicolumn{1}{c|}{84.3}          & \multicolumn{1}{c|}{99.7}          & 7.4          & \multicolumn{1}{c|}{62.8}          & \multicolumn{1}{c|}{96.7}          & 9.6          & \multicolumn{1}{c|}{47.4}          & \multicolumn{1}{c|}{69.4}          & \textbf{3.9} \\ \hline
\textbf{TAPE (Ours)}                 & \multicolumn{1}{c|}{\textbf{51.5}} & \multicolumn{1}{c|}{\textbf{79.9}} & \multicolumn{1}{c|}{\textbf{98.1}} & 8.9         & \multicolumn{1}{c|}{\textbf{59.1}} & \multicolumn{1}{c|}{\textbf{94.2}} & 11.6         & \multicolumn{1}{c|}{\textbf{42.1}} & \multicolumn{1}{c|}{\textbf{62.6}} & 6.5         \\ \hline
\end{tabular}
\end{adjustbox}
\caption{Evaluation of state-of-the-art video-based methods on 3DPW, MPI-INF-3DHP, and Human3.6M datasets. Following Choi \etal~\cite{choi2020beyond}, 
all methods are trained on the training set including 3DPW, but do not use the Human3.6M SMPL parameters obtained with Mosh \cite{Loper:SIGASIA:2014}. The number of input frames follows the original protocol of each method.}
\label{w3dpw}
\end{table*}

\begin{figure}[t]
    \centering
    \includegraphics[width=0.7\textwidth]{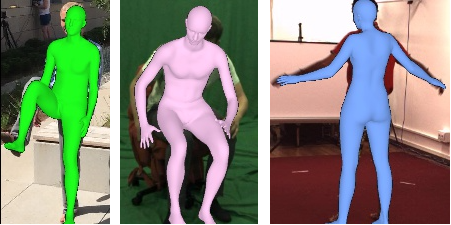}
    \caption{Human body estimation with TAPE in the \textit{first} scenario. Results are shown for 3DPW (left-green), MPI-INF-3DHP (center-pink), Human3.6M (right-blue).}
    \label{qual_alldatasets}
\end{figure}
\subsection{ Comparison with state-of-the-art methods}

We consider two scenarios for assessing the performance of the proposed method. In the \textit{first} scenario, we train our network using MPI-INF-3DHP and Human3.6M augmented with SMPL annotations. We evaluate our method on all datasets (MPI-INF-3DHP, Human3.6M, 3DPW). Evaluating on 3DPW shows how our method generalizes to unknown data and how it estimates human shape and pose in videos with outdoor activities. In the \textit{second} scenario, we, additionally, use  the 3DPW~\cite{vonMarcard2018} dataset  with outdoor scenes during training, but ignore the SMPL annotations in Human3.6M. We evaluate our method again on all datasets. AMASS~\cite{AMASS:ICCV:2019} is used for adversarial training to obtain real samples of 3D human motion in both scenarios.

\begin{table*}[t]
\centering
\begin{adjustbox}{width={0.7\textwidth},totalheight={\textheight},keepaspectratio}%

 \centering

\begin{tabular}{|l|cc|cc|cc|}
\hline
        & \multicolumn{2}{c|}{n=5} & \multicolumn{2}{c|}{n=10} & \multicolumn{2}{c|}{n=25} \\ \hline
        & 3DPW        & H36M       & 3DPW        & H36M        & 3DPW        & H36M        \\ \hline
Biggs   & 57.1        & 42.0       & 56.6        & 42.2        & 55.6        & 42.2         \\
ProHMR  & 56.5        & 39.4       & 54.6        & 38.3        & 52.6        & 36.8         \\
TAPE & \textbf{53.9}        & \textbf{38.1}       & \textbf{52.0}        & \textbf{37.0}        & \textbf{49.5}          & \textbf{35.4}         
\\ \hline
\end{tabular}
\end{adjustbox}

\caption{Multiple hypotheses evaluation. Numbers are PA-MPJPE in mm. We report the minimum error over $n$ samples drawn from the distribution.}
\label{samples}
\end{table*}

\begin{table}[t]
\begin{adjustbox}{width={\textwidth}}%
\begin{tabular}{l|c|c|c|c|c|c|c|}
\cline{2-8}
                                  & \multicolumn{1}{l|}{SPIN} & \multicolumn{1}{l|}{SPIN+} & \multicolumn{1}{l|}{ProHMR} & \multicolumn{1}{l|}{ProHMR+} & \multicolumn{1}{l|}{TAPE} & \multicolumn{1}{l|}{TAPE+} & \multicolumn{1}{l|}{TAPE+} \\ 

                                  & \multicolumn{1}{l|}{ } & \multicolumn{1}{l|}{ SMPLify} & \multicolumn{1}{l|}{ } & \multicolumn{1}{l|}{ fitting} & \multicolumn{1}{l|}{ } & \multicolumn{1}{l|}{SMPLify} & \multicolumn{1}{l|}{fitting} \\ \hline
                                  
\multicolumn{1}{|l|}{PA-MPJPE}    & 41.8                      & 43.8                              & 41.2                        & 34.8                                  & 39.5                      & 38.3                                & \textbf{32.8}                                \\ \hline
\multicolumn{1}{|l|}{ACCEL-ERR} & -                         & -                                 & -                           & -                                     & 6.5                       & 6.2                                 & \textbf{5.4}                                  \\ \hline
\end{tabular}
\end{adjustbox}
\caption{Evaluation of different model fitting methods.
The fitting algorithms are initialized by the corresponding
regression results. All numbers are PA-MPJPE in mm.}
\label{fit}
\end{table}

\vspace*{0.1cm}\noindent\textbf{Video-based methods}: Results in Table~\ref{wo3dpw} show a comparison of our method with the state-of-the-art methods on 3DPW, MPI-INF-3DHP and Human 3.6M for the \textit{first} scenario. Our method outperforms in PA-MPJPE and MPJPE every other method in Human3.6M and MPI-INF-3DHP. Testing on 3DPW without the use of the dataset during training shows how accurately the presented methods can generalize to outdoor videos. In 3DPW, our method produces comparable PA-MPJPE with the state-of-the-art and outperforms them in MPJPE. \\

 In Table~\ref{w3dpw}, we see the results for the \textit{second} training scenario. Ignoring the SMPL annotations in Human3.6M during training drops the performance slightly for all methods. However, our method outperforms all other methods in the PA-MPJPE, MPJPE and MPVPE metrics for all datasets, which is also facilitated by the fact that 3DPW has been added in the training set. These metrics show that our method predicts better poses, body orientation and human shape. 

In both scenarios (Tables~\ref{wo3dpw},~\ref{w3dpw}), the acceleration error of our method is comparable to MPS-NET.

\begin{figure}[t]
    \centering
    \includegraphics[width=0.7\textwidth]{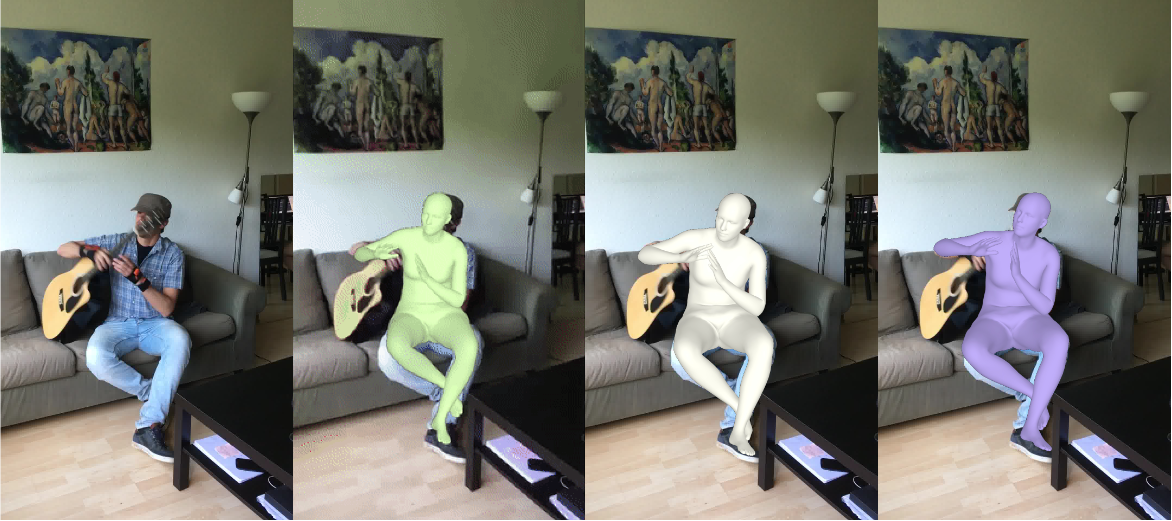}
        \includegraphics[width=0.7\textwidth]{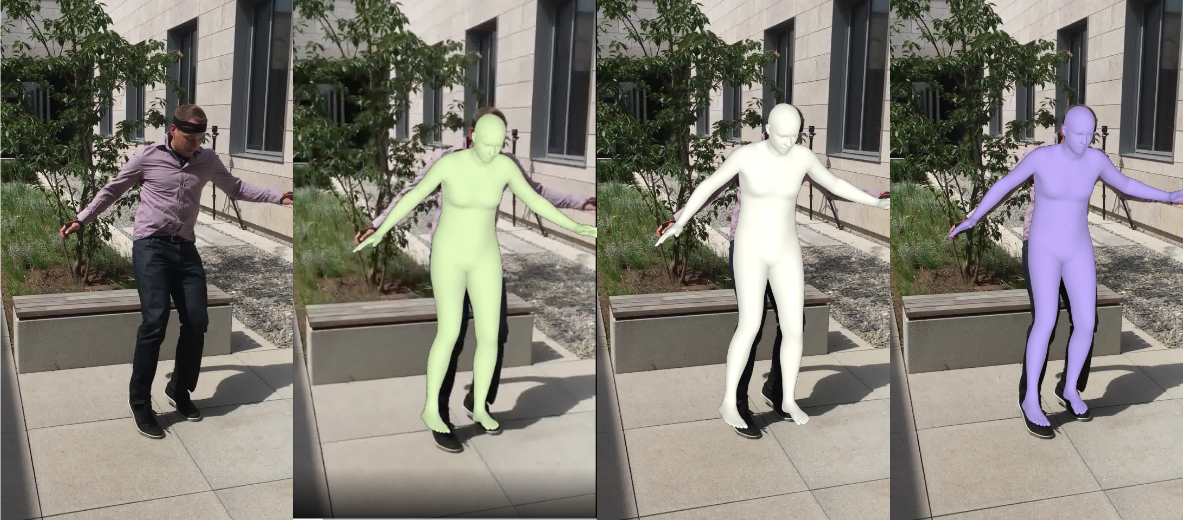}
    \caption{Qualitative results on 3DPW test set~\cite{vonMarcard2018}. From left to right: original input, MPS-Net~\cite{WeiLin2022mpsnet} (light-green), TAPE (white), TAPE + fitting on OpenPose~\cite{wei2016cpm} detections (purple).}
    \label{fig:my_label2}
\end{figure}

\vspace*{0.1cm}\noindent\textbf{Single frame methods}:
Our approach outperforms single frame methods (SPIN, Biggs, ProHMR) and this shows that we effectively capture temporal information from the video input.

\vspace*{0.1cm}\noindent\textbf{Multiple hypotheses}:
In Table~\ref{samples}, we compare the representational power of TAPE with Biggs~\cite{biggs2020multibodies} and ProHMR~\cite{kolotouros2021prohmr} that provide non-deterministic outputs for different number of random samples drawn from the distribution. We consider $5$, $10$ and $25$ samples and report the minimum PA-MPJPE out of all selected samples. Our method outperforms both previous work methods for every number of samples.

\begin{figure}[t]
    \centering
    \includegraphics[width=0.6\textwidth]{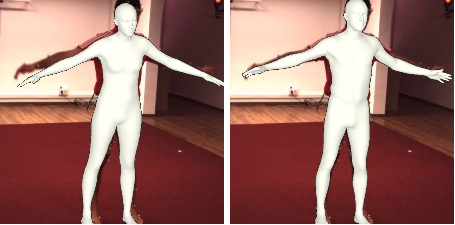}
    \caption{(Left) Human body estimation using TAPE on Human3.6M. (Right) Refinement of TAPE prediction using the proposed optimization framework.}
    \label{opt}
\end{figure}

\vspace*{0.1cm}\noindent\textbf{Optimization task}:
In Table~\ref{fit}, we compare SPIN\cite{kolotouros2019cmr}, ProHMR\cite{kolotouros2021prohmr} and TAPE in conjunction with  optimization-based model fitting frameworks considering 2D keypoints. Evaluation is performed on the Human3.6M dataset. In all cases, optimization is initialized using the regression output of the neural network in each method. The output of SPIN is fitted on 2D keypoints using SMPLify~\cite{Bogo:ECCV:2016}. The output of ProHMR is used in the optimization proposed in the corresponding paper. The output of TAPE is used in the proposed optimization framework. We observe that our proposed optimization framework drops noticeably the already reduced PA-MPJPE of TAPE and is more effective than the widely used SMPLify framework. The acceleration error is improved as well.

\section{Qualitative Evaluation}

\begin{figure}[t]
    \centering
    \includegraphics[width=0.8\textwidth]{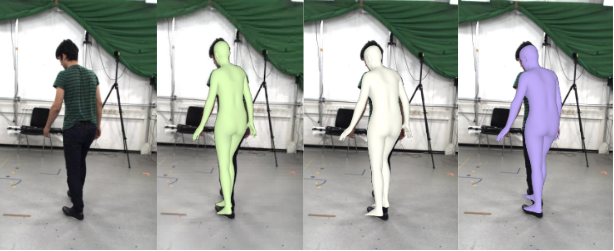}
    \caption{Qualitative on MPI-INF-3DHP test set~\cite{mono-3dhp2017}. From left to right: original input, MPS-Net~\cite{WeiLin2022mpsnet} (light-green), TAPE (white), TAPE + fitting on OpenPose~\cite{wei2016cpm} detections (purple).}
    \label{fig:my_label4}
\end{figure}

We provide qualitative results on the performance of our method against previous work and on various datasets. In Figure~\ref{mps-TAPE}, we show a comparison with the SOTA method MPS-Net on the challenging dataset 3DPW without using the dataset in training (\textit{first} scenario). We observe that our method produces more accurate pose predictions than the current SOTA. 
Figure~\ref{opt} shows the visual impact of the proposed optimization framework at refining the output of TAPE. It is clear that the fitting optimization improves the pose and shape accuracy. 
Optimization on both ground-truth and open-pose keypoints improves the prediction as shown in Figure~\ref{fig:my_label2} on 3DPW dataset and in Figure~\ref{fig:my_label4} on MPI-INF-3DHP.
Finally, Figure~\ref{qual_alldatasets} shows examples of the performance of TAPE on all datasets based on the same scenario.

\section{Conclusions}
We propose TAPE, the first probabilistic temporal model for human pose, shape estimation and camera prediction from video input. We combine an attention-based temporal encoder with a probabilistic model based on normalizing flows and show increased accuracy compared to state-of-the-art and real-time performance. We, additionally, show that optimizing for human pose and shape estimation using TAPE as a video-based prior for human pose outperforms the widely used SMPLify method for 2D keypoint fitting that is image-agnostic by a large margin. Future work includes extending our method to temporal human pose estimation from multiple views and experimenting with Transformer-based temporal encoders to further increase the 3D shape and pose estimation accuracy.

\section*{Acknowledgements}
This work is partially supported by the Greek Secretariat for Research and Innovation and the EU, Project SignGuide: Automated Museum Guidance using Sign Language T2EDK-00982 within the
framework of ``Competitiveness, Entrepreneurship and Innovation" (EPAnEK) Operational Programme 2014-2020. It was also partially supported by the Hellenic Foundation for Research and Innovation (HFRI)  under the ``1st Call for HFRI Research Projects to support Faculty members and Researchers and the procurement of high-cost research equipment'', project I.C.Humans, number 91.
%
%
%

\bibliographystyle{splncs04}
%

\bibliography{egbib}

\end{document}